\def\year{2017}\relax
\documentclass[letterpaper]{article}
\usepackage{nonamestyle}
\usepackage{times}
\usepackage{helvet}
\usepackage{courier}

\usepackage{graphicx}
\usepackage{url}
\usepackage{amsmath}
\usepackage{textcomp}

\begin{document}
%
\title{Information Extraction with Character-level\\Neural Networks and Free Noisy Supervision}
\author{
Philipp Meerkamp\\
  Bloomberg LP\\
  731 Lexington Avenue\\
  New York, NY 10022, USA\\
  pmeerkamp@bloomberg.net\\
  \And
  Zhengyi Zhou \\
  AT\&T Labs Research \\
  33 Thomas Street\\
  New York, NY 10007, USA\\
  zzhou@research.att.com
}
\maketitle

\begin{abstract}
We present an architecture for information extraction from text that augments an existing parser with a character-level neural network. The network is trained using a measure of consistency of extracted data with existing databases as a form of noisy supervision. 
Our architecture combines the ability of constraint-based information extraction systems to easily incorporate domain knowledge and constraints with the ability of deep neural networks to leverage large amounts of data to learn complex features. 
Boosting the existing parser's precision, the system led to large improvements over a mature and highly tuned constraint-based production information extraction system used at Bloomberg for financial language text. 
\end{abstract}

\section{Introduction}

\subsection{Information extraction in finance}

\begin{figure}
\centering
\includegraphics[scale=0.5]{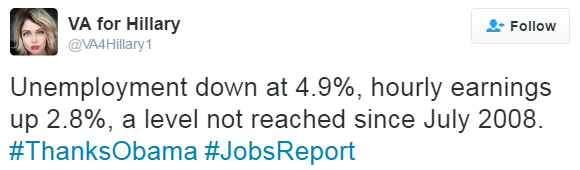}
\caption{Tweet containing economic data.  
Language in the financial domain often trades grammatical correctness for brevity. A multitude of numeric tokens need to be disambiguated into entities such as prices, percentage changes, dates, times, and others. With such a large universe of potential labels, the risk of extracting false positives is high.  }
\label{fig:financial_tweet}
\end{figure}

Unstructured textual data is abundant in the financial domain (see e.g. Figure~\ref{fig:financial_tweet}). This information is by definition not in a format that lends itself to immediate processing. Hence, information extraction is an essential step in business applications that require fast, accurate, and low-cost information processing. In the financial domain, these applications include the creation of time series databases for macroeconomic forecasting or financial analysis, as well as the real-time extraction of time series data to inform algorithmic trading strategies. 
Bloomberg has had information extraction systems for financial language text for nearly a decade. 

To meet the application domain's high accuracy requirements, marrying constraints with statistical models is often beneficial, see e.g. \cite{joint_semantic_role_labeling,constrained_conditional_model}. 
Many quantities appearing in information extraction problems are by definition constrained in the numerical values they can assume (e.g. unemployment numbers cannot be negative numbers, while changes in unemployment numbers \emph{can} be negative). The inclusion of such constraints may significantly boost data efficiency. 
Constraints can be complex in nature, and may involve multiple entities belonging to an extraction candidate generated by the parser. 
At Bloomberg, we found the system for information extraction described in this paper especially useful to extract time series (TS) data. As an example, consider numerical relations of the form {\center \emph{ts\_tick\_abs (TS symbol, numerical value)}, \par}
e.g. \emph{ts\_tick\_abs (US\_Unemployment, 4.9\%)}, or
{\center \emph{ts\_tick\_rel (TS symbol, change in num. value)}, \par}
e.g. \emph{ts\_tick\_abs (US\_Unemployment, -0.2\%)}.\\

\subsection{Our contribution}

We present an information extraction architecture that augments a candidate-generating parser with a deep neural network. The candidate-generating parser may leverage constraints. At the same time, the architecture gains the neural networks's ability to leverage large amounts of data to learn complex features that are tuned for the application at hand. 
Our method assumes the existence of a potentially noisy source of supervision $\Sigma$, e.g. via consistency checks of extracted data against existing databases, or via human interaction. This supervision is used to train the neural network. 

Our extraction system has three advantages over earlier work on information extraction with deep neural networks \cite{CVG,relation_extraction1,relation_extraction2,bidirectional_rnn,end_to_end_sem_role_labeling_recurrent,bidirectional_lstm_cnn,char_dependency,miwa2016relation}:
\begin{itemize}
\item Our system leverages ``free'' data to train a deep neural network, and does not require large-scale manual annotation. 
    The network is trained with noisy supervision provided by measures of consistency with existing databases (e.g. an extraction \emph{ts\_tick\_abs (US\_Unemployment, 49\%)} would be implausible given recent US employment history). With slight modifications, our pipeline could be trained with supervision from human interaction, such as clicks on online advertisements. Learning without explicit annotations is critical in applications where large-scale manual annotation would be prohibitively expensive. 
\item If an extractor for the given application has already been built, the neural network boosts its accuracy without the need to re-engineer or discard the existing solution. 
    Even for new systems, the decoupling of candidate-generation and the neural network offers advantages: the candidate-generating parser can easily enforce contraints that would be difficult to support in an algorithm relying entirely on a neural network. 
    Note that, in particular, a carefully engineered candidate-generating parser enforces constraints intelligently, and can in many instances eliminate the need to evaluate computationally expensive constraints, e.g. API calls. 
\item We encode the candidate-generating parser's document annotations character-by-character into vectors $f_i$ that also include a one-hot encoding of the character itself. We believe that this encoding makes it easier for the network to learn character-level characteristics of the entities in the semantic relation. Moreover, our encoding lends itself well to processing both by recurrent architectures (processing character-by-character input vectors $f_i$) and convolutional architectures (performing $1$D convolutions over an input matrix whose columns are vectors $f_i$).  
\end{itemize}

In a production setting, the neural architecture presented here reduced the number of false positive extractions in financial information extraction application by $>90\%$ relative to a mature system developed over the course of several years.

\section{Design}

\subsection{Overview}

\begin{figure*}
\centering
\hspace{-2mm}
\includegraphics[scale=0.40]{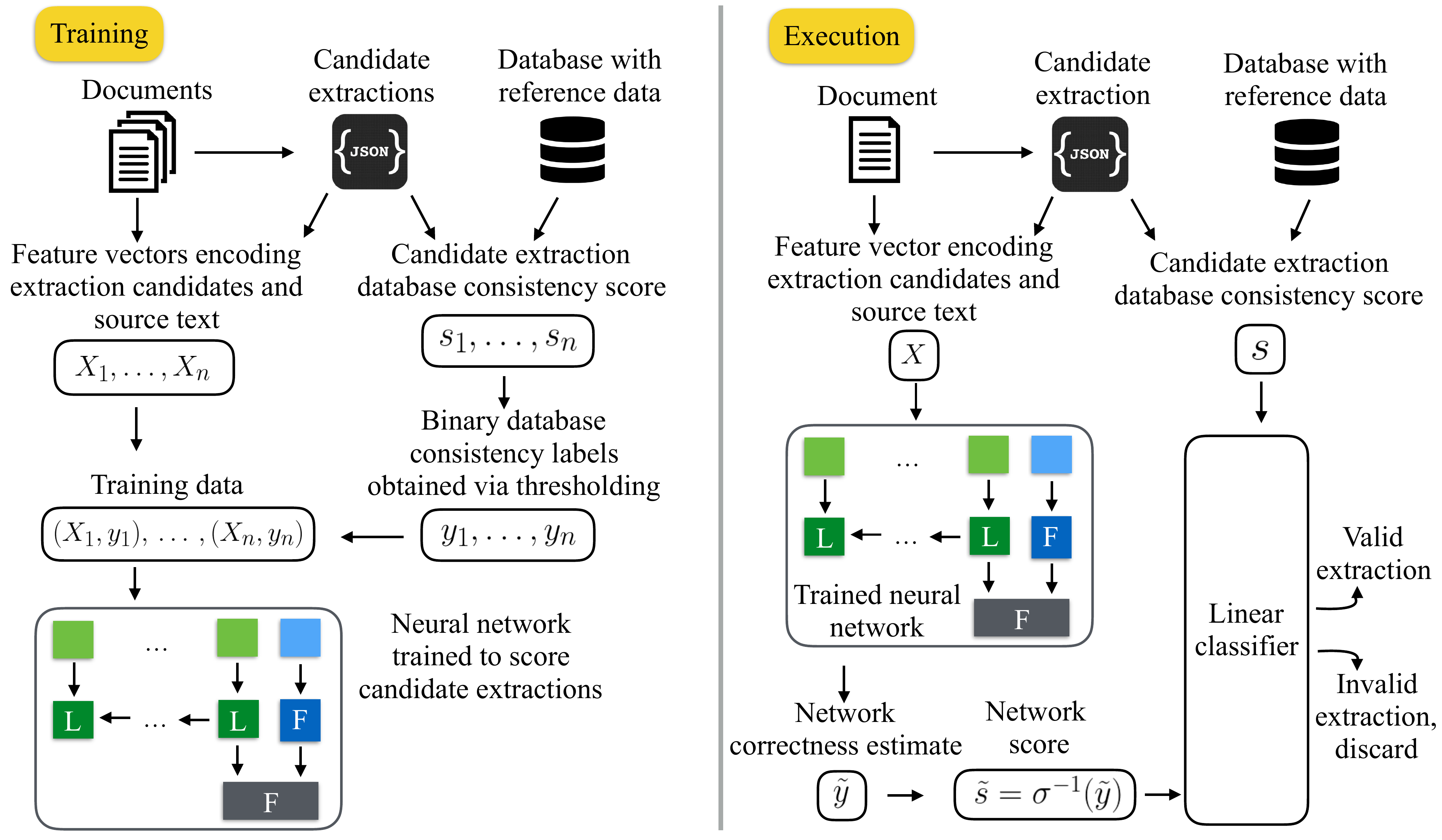}
\caption{Training set-up (left) and execution (right). Blocks marked ``L'' are neural network LSTM cells, while blocks marked ``F'' are fully connected layers. }
\label{fig:training_execution}
\end{figure*}

The information extraction pipeline we developed consists of four stages (see right pane of Figure~\ref{fig:training_execution}). 
\begin{enumerate}
\item The document is parsed using a potentially constraint-based parser, which outputs a set of candidate extractions. Each candidate extraction consists of the character offsets of all extracted constituent entities, as well as a representation of the extracted relation. 
    It may additionally contain auxilliary information that the parser may have generated, such as part of speech tags. 
\item We compute a consistency score $s$ for the candidate extraction, measuring if the extracted relation is consistent with (noisy) supervision $\Sigma$ (e.g. an existing database). 
\item Each candidate extraction generated, together with the section of the document it was found in, is encoded into feature data $X$. A deep neural network is used to compute a neural network candidate correctness score $\tilde{s}$ for each extraction candidate. 
\item A linear classifier classifies extraction candidates as correct and incorrect extractions, based on consistency and correctness scores $s$ and $\tilde{s}$ and potentially other features. Candidates classified as incorrect are discarded.
\end{enumerate}

\subsection{Neural network input and architecture}

\begin{figure*}
\centering
\includegraphics[scale=0.40]{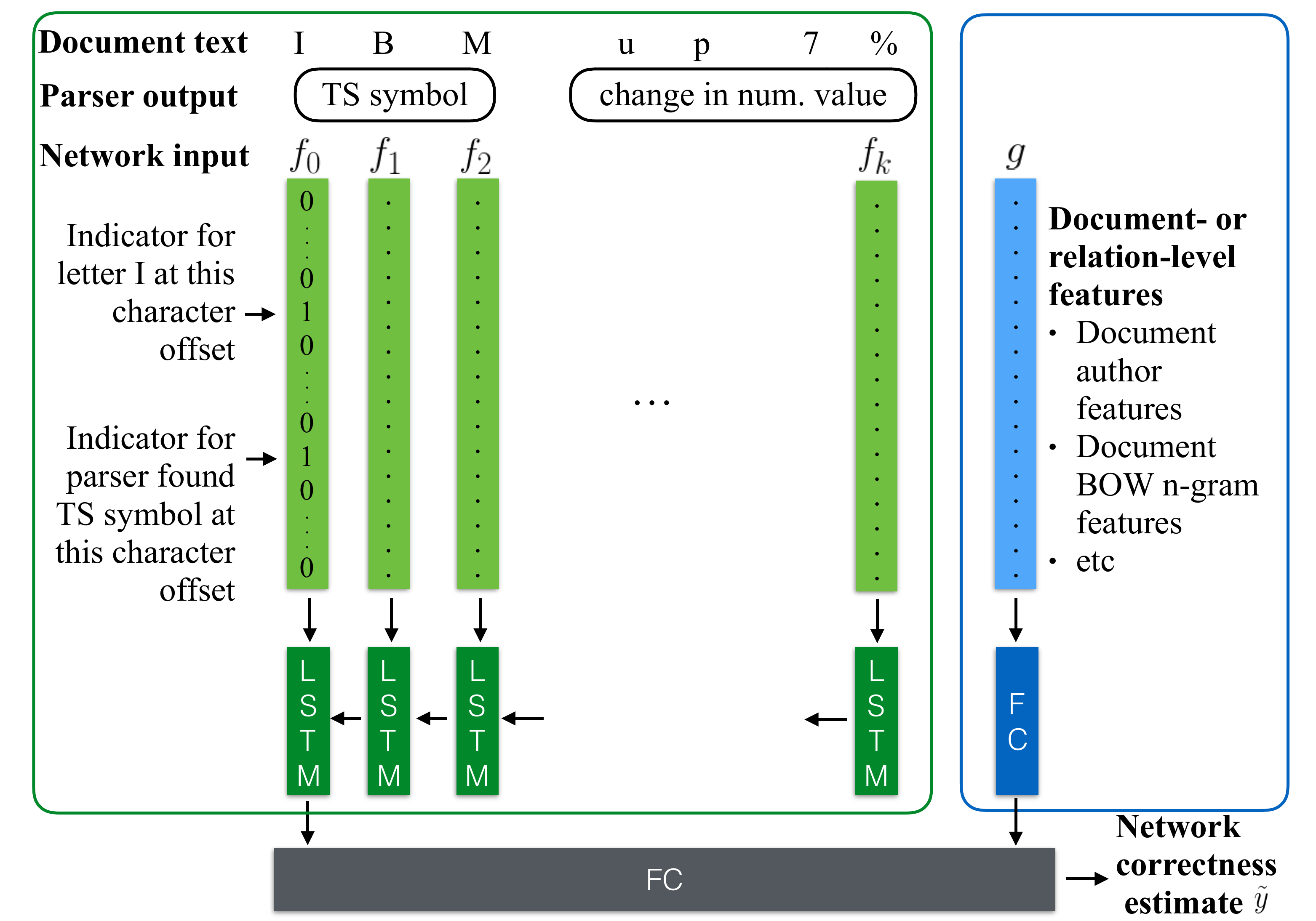}
\caption{
    We use a neural network comprised of an LSTM \cite{lstm}, which processes encoded parser output and a section of document text character-by-character, a fully connected layer (FC, blue) that takes document-level features as input, and a fully connected layer (FC, grey) that takes the output vectors of the previous two layers as input to generate a correctness estimate for the extraction candidate. The final fully connected layer uses a sigmoid activation function to generate a correctness estimate $\tilde{y}\in (0,1)$, from which we compute the network correctness score as $\tilde{s}:=\sigma^{-1}(\tilde{y})$. 
}
\label{fig:input_nn}
\end{figure*}

The neural network processes each input candidate independently. To estimate the correctness of a extracted candidate, the network is provided with two pieces of input (see Figure~\ref{fig:input_nn} for the full structure of the neural network): first, the network is provided with a vector $g$ containing global features, such as attributes of the document's author, or word-level n-gram features of the document text. The second piece of input data consists of a sequence of vectors $f_i$, encoding the document text and the parser's output at a character level. There is one vector $f_i$ for each character $c_i$ of the document section where the extraction candidate was found. 

The vectors $f_i$ are a concatenation of (i) a one-hot encoding of the character and (ii) information about entities the parser identified at the position of $c_i$. For (i) we use a restricted character set of size~$94$, including \texttt{[a-zA-Z0-9]} and several whitespace and special characters, plus an indicator to represent characters not present in our restricted character set. For (ii), $f_i$ contains data representing the parser's output. For our application, we include in $f_i$ a vector of indicators specifying whether or not any of the entities appearing in the relations supported by the parser were found in the position of character $c_i$. 

\subsection{Training and database supervision}

We propose to train the neural network by referencing candidates extracted by a high-recall candidate-generating parser against a potentially noisy reference source (see Figure~\ref{fig:training_execution}, left panel). 
In our application, this reference was a database containing historical time series data, which enabled us to check how well the extracted numerical data fit into time series in the database. 
Concretely, we compute a consistency score $s\in (-\infty, \infty)$ that measures the degree of consistency with the database. 
Depending on the application, the score may for instance be a squared relative error, an absolute error, or a more complex error function. 
In many applications, the score $s$ will be noisy (see below for further discussion). We threshold $s$ to obtain binary correctness labels $y\in\{0,1\}$.
We then use the binary correctness labels $y$ for supervised neural network training, with binary cross-entropy loss as the loss function. 
This allows us to train a network that can compute a pseudo-likelihood $\tilde{y}\in (0,1)$ of a given extraction candidate to agree with the database. Thus, $\tilde{y}$ estimates how likely the extraction candidate is correct.

We assume that the noise in the source of supervision $\Sigma$ is limited in magnitude, e.g. $<5\%$. 
We moreover assume that there are no strong patterns in the distribution of the noise: if the noise correlates with certain attributes of the candidate-extraction, the pseudo-likelihoods $\tilde{y}$ might no longer be a good estimate of the candidate extraction's probability of being a correct extraction. 

There are two sources of noise in our application's database supervision. First, there is a high rate of false positives. It is not rare for the parser to generate an extraction candidate \emph{ts\_tick\_abs (TS symbol, numerical value)} in which the numerical value fits into the time series of the time series symbol, but the extraction is nonetheless incorrect.
False negatives are also a problem: many financial time series are sparse and are rarely observed. As a result, it is common for differences between reference numerical values and extracted numerical values to be large even for correct extractions. 

The neural network's training data consists of candidates generated by the candidate-generating parser, and noisy binary consistency labels $y$.

\section{Results}

The full pipeline, deployed in a production setting, resulted in a reduction in false positives of more than $90\%$ in the extractions produced by our pipeline. The drop in recall relative to the production system was smaller than $1\%$. 

We found that even with only 256 hidden LSTM cells, the neural network described in the previous section significantly outperformed a 2-layer fully connected network with n-grams based on document text and parser annotations as input.

\section{Conclusion}

We presented an architecture for information extraction from text using a combination of an existing parser and a deep neural network. The architecture can boost the precision of a high-recall information extraction system. To train the neural network, we use measures of consistency between extracted data and existing databases as a form of noisy supervision. 
The architecture resulted in substantial improvements over a mature and highly tuned constraint-based information extraction system for financial language text. 
While we used time series databases to derive measures of consistency for candidate extractions, our set-up can easily be applied to a variety of other information extraction tasks for which potentially noisy reference data is available.

\section{Acknowledgments}

We would like to thank my managers Alex Bozic, Tim Phelan and Joshwini Pereira for supporting this project, as well as David Rosenberg from the CTO's office for providing access to GPU infrastructure.

\nocite{*}

\bibliographystyle{aaai}
\bibliography{bibliography}

\end{document}